\documentclass{article}
\usepackage{aux/mlspconf}
\usepackage{amsmath,graphicx}
\usepackage{amssymb}
\usepackage{amsfonts}
\usepackage{multirow}
\usepackage{graphicx}
\usepackage{adjustbox}
\toappear{2024 IEEE International Workshop on Machine Learning for Signal Processing, Sept.\ 22--25, 2024, London, UK}

\title{HIERVAR: A Hierarchical Feature Selection Method for Time Series Analysis}
\name{Alireza Keshavarzian$^1$,  Shahrokh Valaee$^1$}
\address{$^1$Department of Electrical \& Computer Engineering, University of Toronto, Toronto, ON, Canada \\
\normalsize{\textit{alireza.keshavarzian@mail.utoronto.ca, valaee@ece.utoronto.ca}}}

\begin{document}

\maketitle

\begin{abstract}
Time series classification stands as a pivotal and intricate challenge across various domains, including finance, healthcare, and industrial systems. In contemporary research, there has been a notable upsurge in exploring feature extraction through random sampling. Unlike deep convolutional networks, these methods sidestep elaborate training procedures, yet they often necessitate generating a surplus of features to comprehensively encapsulate time series nuances. Consequently, some features may lack relevance to labels or exhibit multi-collinearity with others. In this paper, we propose a novel hierarchical feature selection method aided by ANOVA variance analysis to address this challenge. Through meticulous experimentation, we demonstrate that our method substantially reduces features by over 94\% while preserving accuracy—a significant advancement in the field of time series analysis and feature selection.
\end{abstract}
\vspace{-2mm}
\section{Introduction}
\label{sec:intro}

The escalating utilization of digital tools across various domains has led to a surge in the availability of time-series data, encompassing a wide array of applications such as finance, healthcare, and industrial systems~\cite{keshavarzian2023representation,eldele2021time}. This includes data such as ECG and EEG recordings in healthcare settings, as well as financial market data and sensor readings in industrial contexts. Accurately analyzing such data is crucial for decision-making processes but presents significant challenges. Traditional analysis methods often necessitate specialized expertise for manual interpretation or rely on complex algorithms that lack adaptability across different datasets.

In recent years, there has been a growing interest in feature extraction through random sampling. These techniques offer speed and scalability and are particularly appealing for handling low-sample size high-dimensional training data in the context of time series analysis. Some of these fully randomized representation techniques are Weighted Sums of Random Kitchen Sinks (RKS)~\cite{rahimi2007random,rahimi2008weighted}, (mini)ROCKET~\cite{dempster2020rocket,dempster2021minirocket}, and RASTER~\cite{keshavarzian2023raster} as alternatives to traditional neural networks such as CNNs and RNNs for time series analysis\cite{ismail2020inceptiontime}. These randomized methods offer a departure from the conventional reliance on extensive training, yet manage to achieve comparable or even superior performance across various tasks, including time series classification. However, their effectiveness comes with a caveat—they demand many features to attain high accuracy, posing challenges for implementation on resource-constrained devices. 
Despite the comparable performance of random representation techniques with deep learning methods, their intrinsic nature of data-independent feature extraction often leads to a significant portion of redundant features. This redundancy can contribute to issues such as overfitting, decreased interpretability, and longer evaluation times.

Recent studies tried to tackle this issue by introducing feature selection designed for variants of the ROCKET method. S-ROCKET~\cite{salehinejad2022s} investigates best features by adding random features through an iterative process without losing accuracy. E-ROCKET~\cite{omidi2023reducing} uses knee/elbow detection on the sorted curve of the L2 regression coefficient vector and chooses coefficients with higher magnitudes. E-ROCKET performs better both in terms of accuracy and computations.
Also, there are other types of feature selection methods such as using Least Absolute Shrinkage and Selection Operator (LASSO), which are mostly used to shrink redundant features toward zero. However, LASSO struggles with correlated predictors, as it tends to arbitrarily select one among them and set the coefficients of others to zero~\cite{freijeiro2022critical}. 
LARS (Least Angle Regression) is another technique  used for feature selection and regression tasks, particularly effective with multicollinearity. It iteratively selects predictors most correlated with the response variable and incrementally adjusts their coefficients towards least square estimates until another predictor becomes equally correlated with the residual, efficiently navigating the feature space. while LARS efficiently handles situations with a large number of predictors, its performance may degrade when the number of predictors greatly exceeds the number of observations. Moreover, LARS can be computationally intensive for very large datasets due to its iterative nature~\cite{freijeiro2022critical}.

In this paper, we introduce a hierarchical feature selection method termed HEIRVAR. By integrating the ANOVA algorithm~\cite{lowry2014concepts} with the E-ROCKET technique, we effectively address the challenge of feature redundancy, ensuring that selected features correlate closely with label information. Notably, while we showcase the effectiveness of HEIRVAR with E-ROCKET, it important to highlight its versatility; it can seamlessly integrate with other feature selection methods like LASSO. Our research reveals that the fusion of ANOVA with diverse feature selection approaches not only streamlines feature pruning but also markedly enhances model accuracy.

\section{Problem Definition}
\label{sec:format}

Let us consider a dataset $\mathcal{D}$ comprising $N$ samples represented as \ $\mathcal{D} = \{(x_n, y_n)\}_{n=1}^N$ , where each $(\boldsymbol{x}_n, y_n)$  pair denotes the $n^{\textit{th}}$ sample containing an input  $\boldsymbol{x}_n$  and its corresponding categorical output $y_n$. Here, $\boldsymbol{x}_n$ is an  $L$-dimensional real vector ($\boldsymbol{x}_n \in \mathbb{R}^L$), where $L$ representing the length of time series, and  $y_n$ represents the label assigned to  $\boldsymbol{x}_n$. Each  $y_n$ belongs to the set $\mathcal{C} = \{1, ..., C\}$, where  $C$ signifies the number of classes. Given a training set as described above, we aim to train a classifier to predict the label  $y$ associated with an input  $\boldsymbol{x}$.

\begin{figure}
    \centering
    \includegraphics[width=1\linewidth]{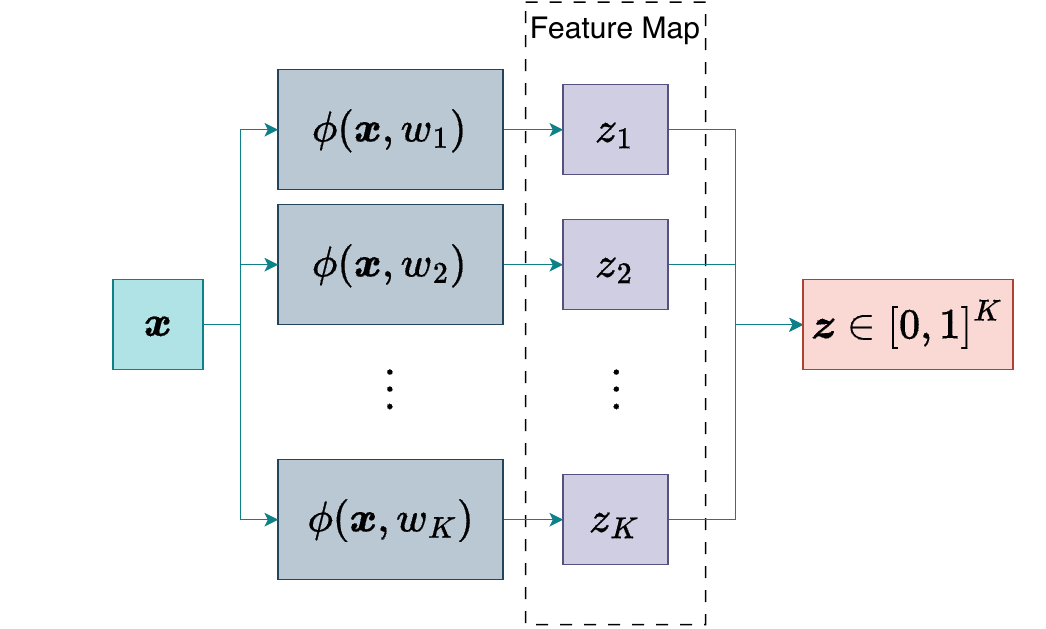}
    \caption{Overall schema of random representation learning}
    \label{fig:representation}
\end{figure}

Here, we present a concise overview of the feature representation process using either ROCKET, MiniROCKET or RASTER. During the training phase, the inputs (denoted as $\boldsymbol{x}_n$) from the training dataset $\mathcal{D}$ undergo transformation via MiniROCKET. This transformation entails convolving each input with $K$ random kernels, followed by non-linear operations, to extract desired features per convolution. The resulting features are acquired through either the Threshold Exceedance Rate (TER) or the Randomized Threshold Exceedance Rate (rTER)~\cite{keshavarzian2023raster}, which quantifies how many times the convolved signal surpasses a random threshold. Consequently, each $\boldsymbol{x}_n$ is converted to $\boldsymbol{z}_n$, where $\boldsymbol{z} \in [0,1]^{K}$ and $K$ denotes the number of features.
Fig. \ref{fig:representation} illustrates the overall architecture of the representation learning process. To maintain simplicity, all convolution, pooling, and additional processing are encapsulated within the function $\phi(.)$, which accepts a vector of hyperparameters $w$.

 Ultimately, the classifier assigns the predicted label $\hat{y}$ to the observation $z$, aiming for a high probability that the predicted label $\hat{y}$ matches the true label $y$. These classifiers can be logistic, L1 or ridge classifiers. 
 
In the data representation process, an extensive array of features is required to capture the full temporal and frequency behaviours of the signal. However, as the number of these features increases, so does the presence of redundant and collinear features. This abundance of features can degrade both the computational efficiency and the performance of the classification task, potentially leading to overfitting and increased processing time. In the next section, we briefly elaborate on the E-ROCKET feature selection mechanism and describe why it is necessary to be combined with another feature selection algorithm. 

\section{E-ROCKET}

E-ROCKET~\cite{omidi2023reducing} operates through three stages: pre-training, knee/elbow detection, and post-training. During the pre-training phase, the signal undergoes transformation via one of the random representation methods discussed earlier, yielding $K$ features for each sample. By constructing the design matrix, denoted as $Z$, which forms an $N \times K$ data matrix, we prepare the data for further processing. Subsequently, the data matrix is inputted into the ridge classification algorithm to compute the corresponding weights for each feature.

\begin{equation}
    W = (Z^{\top}Z + \lambda I )^{-1}Z^{\top}Y, 
\end{equation}
where $Y$ is $N\times C$ matrix, and each row of it is a one-hot vector corresponding to the label of the sample. In the next phase, it will sort the coefficient of ridge classification and choose coefficients that have higher magnitudes by applying the KNEEDLE algorithm~\cite{satopaa2011finding}. The intuition behind the E-ROCKET algorithm is that coefficients that have higher values correspond to the features that have higher contribution and importance to the dependent variable. For simplicity, assume that we are dealing with binary classification. Then, $W$ will be a vector of size $K$ where $K$ indicates the number of features. We denote the sorted version of $W$ by $U$ as follows: 
\begin{equation}
    U = \operatorname{sort}(|W|)
\end{equation}
where $|.|$ is the absolute operator. So we have $U_1 \leq U_2 \leq \cdots \leq U_K$. E-ROCKET applies the KNEEDLE algorithm to find the knee point in the curve of $U$, where the knee is the high-curvature point, $k^{*}$, of a sequence of $U$. Keeping the higher value coefficient and discarding the lower part of the knee point will result in a subset of features that have enough ability to perform the classification accuracy. Assume that after finding the knee point we keep $K'$ features, out of $K$, where $\mathcal{S}$ indicates the set of indices of the selected features. Then, each sample after applying feature selection is denoted by: 
\begin{equation}
    \label{eqn:erocket}
    \boldsymbol{z}'_n = [z_{n,{k|k\in\mathcal{S}}}]
\end{equation}
The transformed and feature-selected dataset can be written as: 
\begin{equation}
    \mathcal{D}^{'} := \{\boldsymbol{z}^{'}_n ,y_n\}^{N}_{n=1},
\end{equation}
Now, we can apply another ridge regression based on the new dataset and perform the classification task. 

\section{ANOVA}
Analysis of Variance (ANOVA)~\cite{satopaa2011finding} is a statistical method used to assess the significance of features in a dataset by comparing the means of multiple groups. In the context of feature selection, ANOVA can be employed to determine which features have a significant impact on the target variable, thereby helping to identify the most informative features for a given problem.

Let $Z = {\boldsymbol{z}_1, \boldsymbol{z}_2, \dots, \boldsymbol{z}_N}$ be a dataset with $N$ instances, where each instance $\boldsymbol{z}_i$ is represented by a vector of $K$ features $\boldsymbol{z}_i = (z_{i1}, z_{i2}, \dots, z_{iK})$. The target variable $y = (y_1, y_2, \dots, y_N)$ corresponds to the class labels or continuous values associated with each instance.

The one-way ANOVA tests the null hypothesis $H_0$ that the means of all groups are equal against the alternative hypothesis $H_1$ that at least one group mean is different from the others. In the context of feature selection, the groups are defined by the unique values of a particular feature $z_{(j)}$, indicating the $j^{\textit{th}}$ feature.

For a given feature $z_{(j)}$, the total sum of squares (SST) can be decomposed into the between-group sum of squares (SSB) and the within-group sum of squares (SSW):

$$
SST=SSB+SSW
$$

The between-group sum of squares is calculated as:

$$
SSB=\sum_{c=1}^C N_c\left(\bar{z}_{(j)}^{(c)}-\bar{z}_{(j)}\right)^2
$$
where $C$ is the number of unique labels for feature $z_{(j)}$, $N_c$ is the number of instances in group $c$, $\bar{z}_{(j)}^{(c)}$ is the mean of feature $z_{(j)}$ for group $c$, and $\bar{z}_{(j)}$ is the overall mean of the feature $z_{(j)}$.
The within-group sum of squares is calculated as:

$$
SSW=\sum_{c=1}^C \sum_{i=1}^{N_c}\left(z_{i j}^{c}-\bar{z}_{j}^{(c)}\right)^2
$$
where $z_{ij}^{c}$ is the value of feature $z_{(j)}$ for the $i^{\textit{th}}$ instance in group $c$.

The F-statistic for feature $z_{(j)}$ is then calculated as:
\begin{equation}
    \label{eqn:ANOVA}
    F_j=\frac{S S B_j /(C-1)}{SSW_j /(N-C)}
\end{equation}
The F-statistic follows an F-distribution with $(C - 1)$ and $(N - C)$ degrees of freedom under the null hypothesis~\cite{ott2016introduction}. A high F-value indicates that the means of the groups are significantly different, suggesting that the feature $z_j$ has a strong influence on the target variable.

To perform feature selection using ANOVA, the F-statistic is calculated for each feature, and the features are ranked based on their F-values. A predefined threshold or a specific number of top-ranked features can be selected for further analysis or model building. However, finding how many features a dataset needs to perform accurately similar to the full feature version is challenging. One way is to evaluate the p-values associated with each of the F-Score of the features; however, that will keep too many features.

\begin{figure}
    \centering
    \includegraphics[width=1\linewidth]{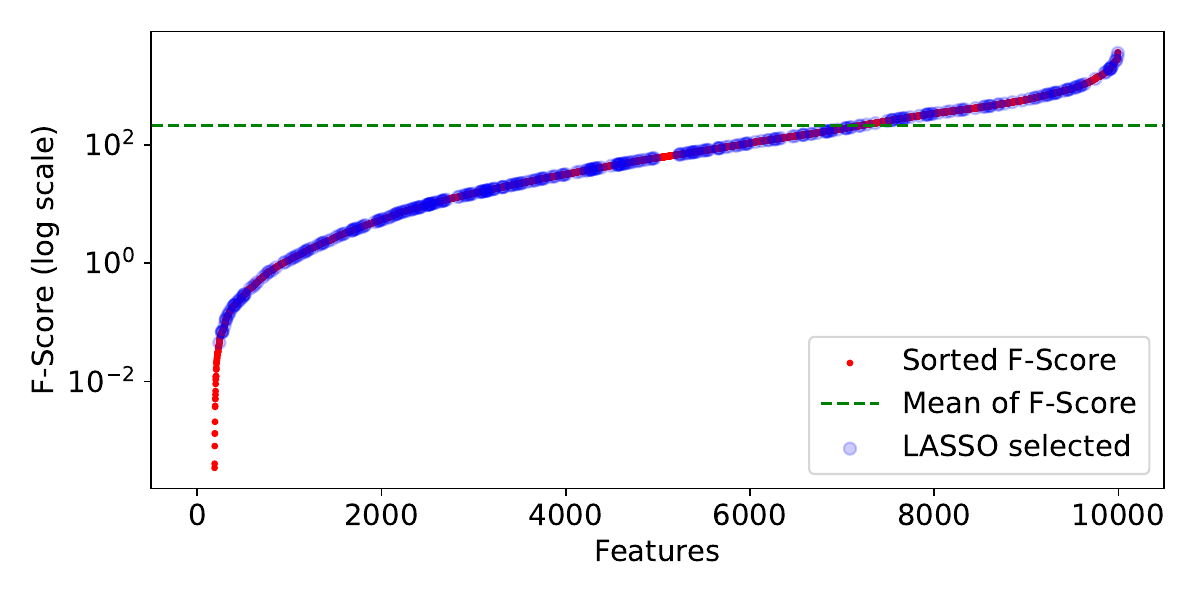}
    \caption{Sorted F-Score of the FordB dataset~\cite{UCRArchive}.}
    \label{fig:sorted}
\end{figure}
\begin{figure*}[ht]
    \centering
    \includegraphics[width=1\linewidth]{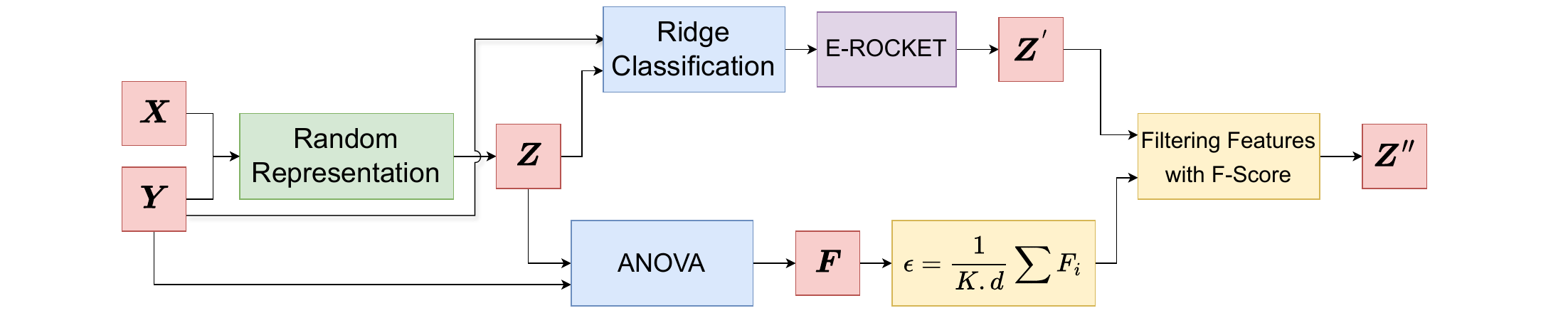}
    \caption{Overall architecture of the proposed method}
    \label{fig:Overallarchitecture}
\end{figure*}
 In this paper, we propose a hierarchical form where feature selection will go through two phases: in the first phase E-ROCKET selects features that cooperate more with the label. In the subsequent phase, we aim to refine our approach by excluding features with lower F-scores identified by the ridge classifier. Our focus is on retaining the most statistically significant and discriminative features, thereby enhancing the model's performance. This can also improve the interpretability of the model, as the selected features are more likely to have a clearer and stronger relationship with the target variable. Moreover, By removing low F-score features from the selected features, we focus on the most statistically significant and discriminative features. This can improve the interpretability of the model, as the selected features are more likely to have a clearer and stronger relationship with the target variable.

The feature set obtained by the E-ROCKET method can be derived by \eqref{eqn:erocket} where the new set indices are denoted by $\mathcal{S}$. In the second phase, we apply the ANOVA algorithm to compute the F-score for each feature in the original feature set $\mathcal{K} := {1, \ldots, K}$. The F-score for the $k$-th feature can be calculated using \eqref{eqn:ANOVA}. So we have a set of F-Score for each feature in the main feature set, $\mathcal{K}$.

To determine the threshold for the features based on their F-scores, we calculate the mean F-score over the original feature set:
\begin{equation}
\mu = \frac{1}{K} \sum_{k=1}^K F_k.
\end{equation}
The threshold $\epsilon$ for passing features with F-scores higher than the mean is derived as:
\begin{equation}
\epsilon = \frac{\mu}{d}.
\end{equation}
Here, $d$ acts as a hyperparameter controlling feature selection strictness in the second phase. A higher $d$ permits more E-ROCKET selected features to pass ANOVA filtering. Fig. \ref{fig:sorted} displays sorted F-Scores from MiniROCKET. Blue dots represent E-ROCKET selected features, with many falling below the threshold, and some near zero.

Finally, the transformed dataset $\mathcal{D}^{"}$ is obtained by selecting the features that satisfy both the E-ROCKET and ANOVA criteria:
\begin{equation}
\boldsymbol{z}^{"}_n = [z_{n,k | k \in \mathcal{S} \land F_k > \epsilon}],
\end{equation}
where $\boldsymbol{z}^{"}_n$ represents the transformed feature vector for the $n$-th instance. Fig. \ref{fig:Overallarchitecture} represents the overall architecture of the proposed method. Note that the Box ``Random Representation" refers to Fig. \ref{fig:representation}.

In summary, removing low F-score features from the ridge classification selected features can be an effective approach to improve the model's performance by leveraging the strengths of both methods, eliminating noisy or irrelevant features, reducing overfitting, and enhancing interpretability.

\begin{table*}[ht]
\centering

\caption{Comparison of Classification Accuracy Before and After Integration of HIERVAR Component in Feature Selection Method, Averaged Over Four Runs over all 108 dataset of UCR archive~\cite{dau2019ucr,UCRArchive}}.

\begin{tabular}{cc|cc|ccc}
\multirow{2}{*}{}         &          & \multicolumn{2}{c}{MiniROCKET} & \multicolumn{2}{c}{RASTER} &  \\ \hline
                          &          & Feature       & Accuracy (\%)       & Feature     & Accuracy (\%)    &  \\ \hline
\multirow{2}{*}{LASSO}    & Original & 1235.1        & 84.92          & 1504.3      & 83.99        &  \\
                          & HIERVAR  & \textbf{461.3}         & \textbf{85.19}          & \textbf{564.7}       & \textbf{84.32}        &  \\ \hline
\multirow{2}{*}{E-ROCKET} & Original & 1511.7        & 85.19          & 1669.1      & 84.81        &  \\
                          & HIERVAR  & \textbf{551.6}         & \textbf{85.27}          & \textbf{613.0}         & \textbf{85.23}        & 
                          \\ \hline
\end{tabular}%
\label{tab:accuracy}

\end{table*}

\section{Computation Complexity}
We have compared the computational complexity of our proposed method with E-ROCKET and LASSO. The pre-training phase where time series undergo the random representation process takes $O(LNK + N^2 K)$ operations which is the computational complexity of MiniROCKET or RASTER. The knee detection algorithm in the E-ROCKET takes $O(K^2)$ operations and after feature selection, there will be another training process on the selected feature set with the complexity $O(LNK' + N^2 K')$. The computational complexity of ANOVA is comprised of calculating the group means, overall means, SSB, and SSW. Group mean, overall mean, and SSW has the complexity $O(N)$ and SSB has the complexity of $O(C)$. Therefore, the overall computational complexity of one-way ANOVA is $O(N + C)$ where $C$ is usually less than $N$. So the overall complexity of the proposed method is written in the form of $O(LNK + N^2 K + N) = O(LNK + N^2 K)$ where it is negligible compared to the previous stages. If we substitute the first stage (E-ROCKET) with other feature selection methods such as LASSO, its computational complexity will be $O(K^3 + K^2 N)$. This makes LASSO more computationally expensive than the Ridge regression for a large number of features. Again using ANOVA, after using LASSO, does not affect the computational complexity.

\section{Experimental Result}

To evaluate the effectiveness of the proposed hierarchical feature selection approach, we conduct experiments on several time series benchmark datasets in the UCR archive~\cite{dau2019ucr,UCRArchive}. 
For the pre-training phase, we use MiniROCKET~\cite{dempster2021minirocket} and RASTER~\cite{keshavarzian2023raster} as two novel and fast random representation learning methods. We set the number of features for both methods as $K = 10,000$. For both MiniROCKET and RASTER, we set the hyperparameters as it was in the original papers~\cite{keshavarzian2023raster,dempster2021minirocket}. For imbalanced datasets, a weighted optimizer is utilized, penalizing errors in minority labels more to enhance model performance. The regularization parameter for the ridge classifier is determined via cross-validation, selecting from the set of values: $\{0.001, 0.01, 0.1, 1\}$.

We compared the performance of models trained on the features selected by our method against models trained on features selected by E-ROCKET~\cite{omidi2023reducing} and LASSO individually. 
E-ROCKET method leverages from cross-validation process to ensure they are using an optimized regularization hyperparameter.

In Table \ref{tab:accuracy}, we present a comparative analysis of two prominent representation learning methodologies, Mini-ROCKET and RASTER, as random representation components. Each method generates 10,000 features with consistent parameters for dilation, thresholding, and padding, as outlined in their respective GitHub repositories. Additionally, we contrast E-ROCKET with LASSO as the initial component of feature selection. In our implementation of LASSO, we specify a regularization parameter, $\alpha = 0.0001$. It's worth noting that adjusting the value of $\alpha$ in LASSO results in selecting fewer features. However, for the sake of comparison with E-ROCKET, we maintain $\alpha$ at 0.0001 to ensure a comparable range of feature selection. 

The findings presented in Table \ref{tab:accuracy} underscore the efficacy of employing LASSO and E-ROCKET as initial feature selection stages. HIERVAR, leveraging approximately 1,000 fewer features compared to the original method, consistently outperforms in accuracy. This improvement holds true across both MiniROCKET and RASTER methodologies, indicating its robustness and superiority in feature selection. 

\begin{figure}
    \centering
    \includegraphics[width=1\linewidth]{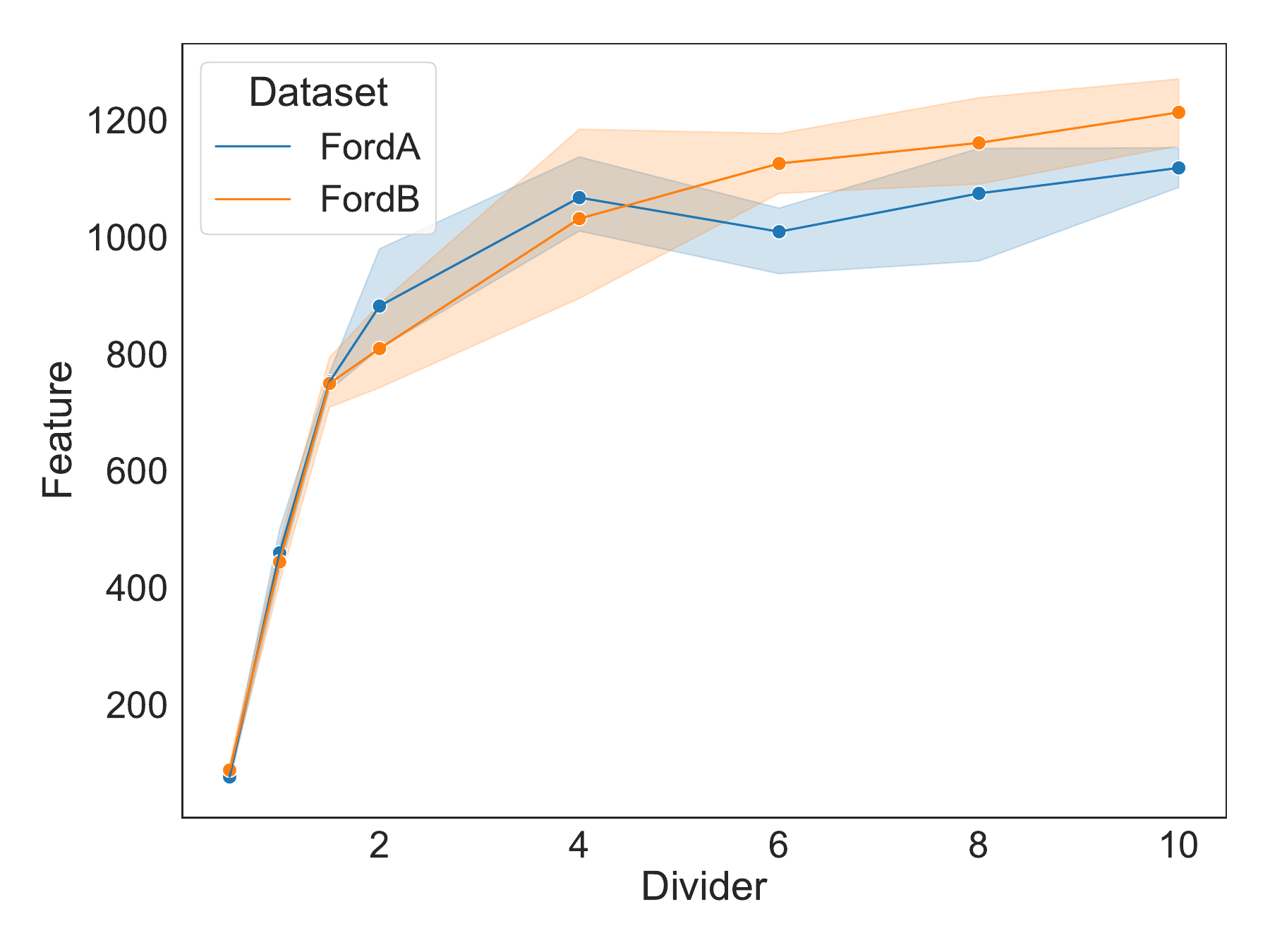}
    \caption{The trend of the number of features as the parameter \(d\) varies.}
    \label{fig:trend}
\end{figure}

Fig. \ref{fig:trend} illustrates the relationship between the number of features and the increment of the divider, \(d\). It demonstrates that as \(d\) increases, the number of features rises initially, reaching a plateau after \(d = 2\), where only marginal increases are observed. It is important to note that this trend is based on analysis using multiple datasets as samples to showcase the behaviour of HIERVAR.

\begin{figure}
    \centering
    \includegraphics[width=1\linewidth]{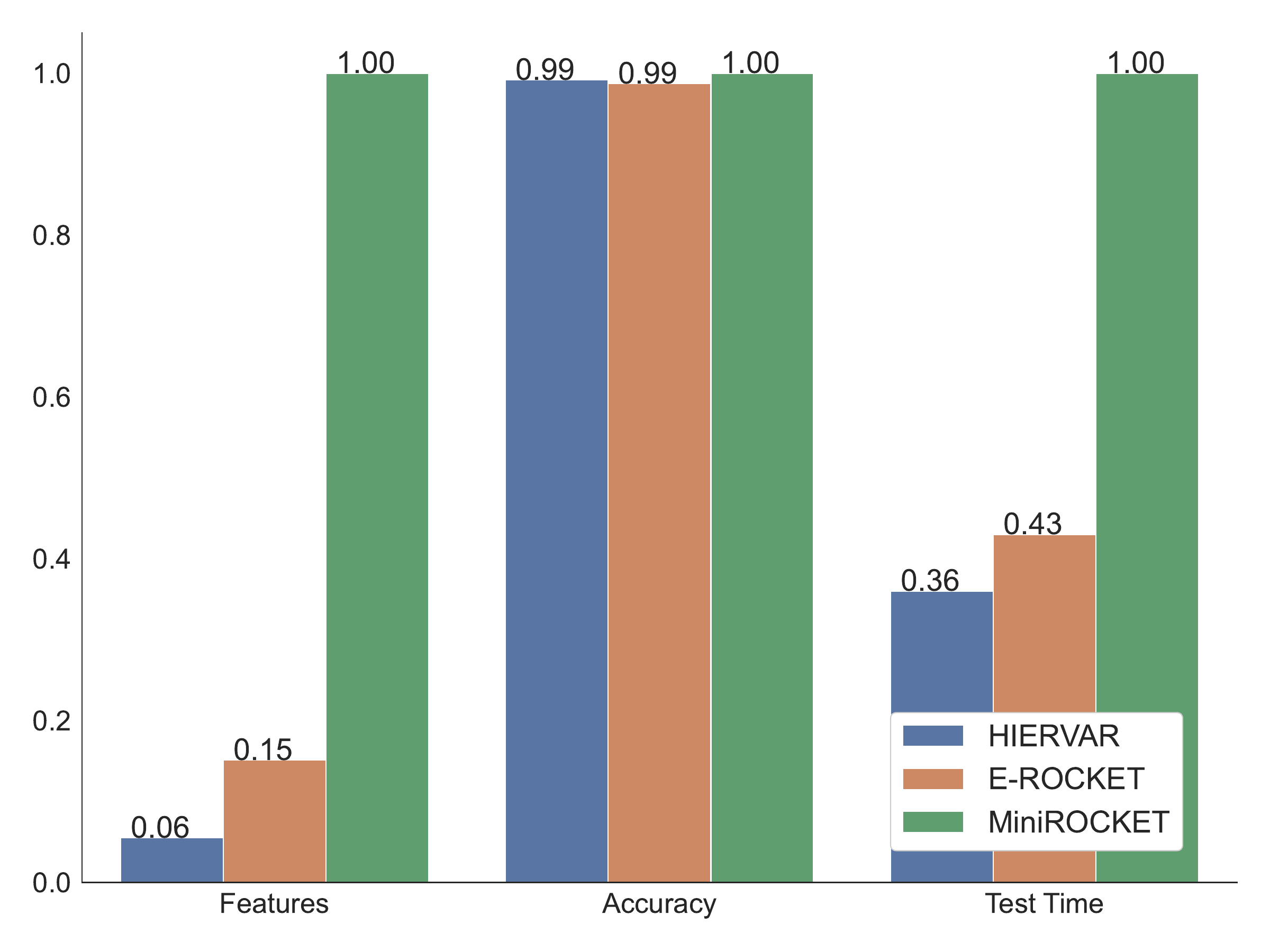}
    \caption{Comparative performance analysis of MINIROCKET, E-ROCKET and HIERVAR. Segmentation is based on the number of features relative to MINIROCKET. Average accuracy is depicted within a discernible range.}
    \label{fig:amin}
\end{figure}

In this experimental analysis, the range of selected features spanned from a minimum of 2 to a maximum of 6105. The observed maximum improvement in accuracy over the baseline MINIROCKET reached 5\%, while the maximum degradation was limited to 4.4\%. Hence, HIERVAR effectively eliminates redundant features, ensuring the preservation of MINIROCKET's classification accuracy. In Fig. \ref{fig:amin}, we present the comprehensive outcome alongside the test duration. As illustrated, employing HEIRVAR over an extended period results in an approximately threefold reduction in total runtime. Consequently, this renders it well-suited for computational-constrained IoT devices.

\section{Conclusion}
This paper introduces a novel hierarchical feature selection method, HIERVAR, aimed at addressing the challenge of feature redundancy in time series classification. By leveraging ANOVA variance analysis, HIERVAR substantially reduces the number of features while maintaining classification accuracy, marking a significant advancement in the field. Our results demonstrate that HIERVAR successfully reduces over 94 percent of random features while consistently outperforming other feature selection techniques in terms of accuracy. 
\vfill
\pagebreak

\bibliographystyle{aux/IEEEbib}
\bibliography{refs}

\end{document}